\pdfoutput=1
\documentclass[11pt]{article}

\usepackage{acl}
\usepackage{verbatim}
\usepackage{bm}
\usepackage{times}
\usepackage{latexsym}
\usepackage{amsmath}
\usepackage{amssymb}
\usepackage[mathscr]{eucal}
\usepackage{mathrsfs}
\usepackage{amsfonts}
\usepackage{mathrsfs} 
\usepackage{multirow} 
\usepackage{booktabs}
\usepackage{amsmath}
\usepackage{multirow}
\usepackage{subcaption}
\usepackage{amssymb,mathrsfs,amsmath}
\usepackage{algorithm}  
\usepackage{algpseudocode}  
\usepackage{amsmath}  
\usepackage{xcolor}
\usepackage{graphicx}
\usepackage{hyperref}
\usepackage[T1]{fontenc}
\usepackage[utf8]{inputenc}
\usepackage{microtype}
\usepackage{inconsolata}
\makeatletter
\renewcommand{\maketag@@@}[1]{\hbox{\m@th\normalsize\normalfont#1}}%
\makeatother

\title{Question-guided Knowledge Graph Re-scoring and Injection for Knowledge Graph Question Answering}

\author{Yu Zhang$^{1}$,Kehai Chen$^{1}$\thanks{Corresponding Author},Xuefeng Bai$^{1}$,Zhao Kang$^{2}$,Quangjiang Guo$^{2}$,Min Zhang$^{1}$ \\
\textsuperscript{1}Institute of Computing and Intelligence, Harbin Institute of Technology, Shenzhen, China \\
\textsuperscript{2}University of Electronic Science and Technology of China, Chengdu, China \\
\texttt{yuzhang2717@gmail.com,\{chenkehai,baixuefeng,zhangmin2021\}@hit.edu.cn,} \\
\texttt{zkang@uestc.edu.cn}, guochance1999@163.com}
\begin{document}
\maketitle

\begin{abstract}
Knowledge graph question answering (KGQA) involves answering natural language questions by leveraging structured information stored in a knowledge graph. 
Typically, KGQA initially retrieve a targeted subgraph from a large-scale knowledge graph, which serves as the basis for reasoning models to address queries. 
However, the retrieved subgraph inevitably brings distraction information for knowledge utilization, impeding the model's ability to perform accurate reasoning. 
To address this issue, we propose a Question-guided Knowledge Graph Re-scoring method (Q-KGR) to eliminate noisy pathways for the input question, thereby focusing specifically on pertinent factual knowledge.
Moreover, we introduce Knowformer, a parameter-efficient method for injecting the re-scored knowledge graph into large language models to enhance their ability to perform factual reasoning.
Extensive experiments on multiple KGQA benchmarks demonstrate the superiority of our method over existing systems. 
\footnote{The code and data are released on \url{https://github.com/EchoDreamer/Q-KGR}.} 
\end{abstract}

\section{Introduction}
Knowledge graph question answering (KGQA) aims to provide factual answers to natural language 
questions by leveraging information from knowledge graphs~\cite{EMKG,structgpt,knowagent}. Sitting at the intersection between graphs and texts, this task can
further facilitate the applicability of knowledge graphs in more downstream tasks, such as recommendation systems~\cite{Personalization,Recommender}, information extraction ~\cite{tpn,FCDS}, time series forecasting~\cite{transformer,token,time}.
\begin{figure}[t]
    \centering
    \includegraphics[page=1, scale=0.58]{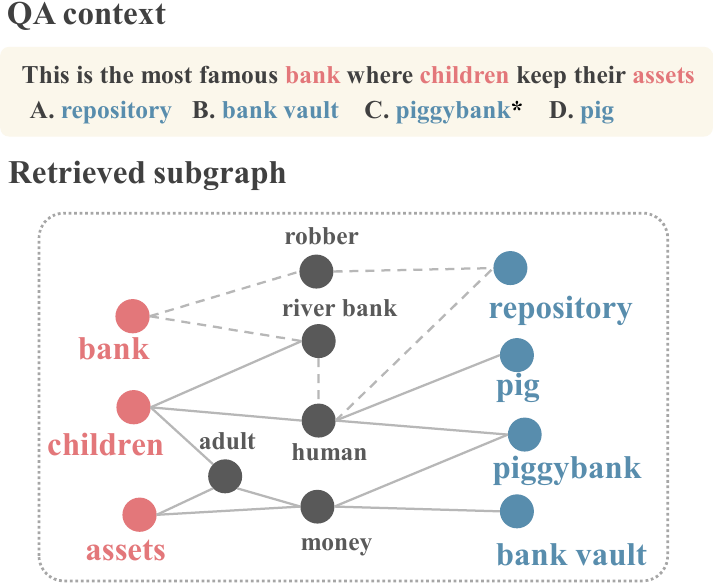}
    \caption{Given the QA context (question and answers), we retrieve a subgraph including question entities (red), answer entities (blue) and other neighbor entities (grey). The correct answer is piggybank.}
    \label{fig:motivation}
    \vspace{-0.5cm}
\end{figure}

Recent work on KGQA attempts to synergize large language models (LLMs) with knowledge graphs (KGs) by utilizing retrieved knowledge from KGs to assist LLMs in conducting factual reasoning (\citealp{rog}; \citealp{knowgpt}; \citealp{tog}). 
A prevailing approach is to heuristically retrieve a subgraph from a large-scale KG using topic entities (KG entities mentioned in the given question and candidate answers) and their few-hop neighbors, and then use this subgraph as the foundation for LLMs to perform reasoning.
Despite the promising performance, this approach  heavily rely on the retrieved graph which inevitably includes distracting information.
This might leads the LLM to focus on irrelevant knowledge, thereby hindering its ability to conduct accurate reasoning.
Figure~\ref{fig:motivation} depicts an example from the RiddleSence dataset \cite{riddle}. 
In the retrieved subgraph, the question entity \textit{\textcolor[RGB]{202,12,22}{bank}} connects a distracting answer entity \textit{\textcolor[RGB]{88,141,173}{repository}} with two noisy pathways: (\uppercase\expandafter{\romannumeral1}) \textit{bank}$\rightarrow$\textit{river bank}$\rightarrow$\textit{human}$\rightarrow$\textit{repository} and (\uppercase\expandafter{\romannumeral2}) \textit{bank}$\rightarrow$\textit{robber}$\rightarrow$\textit{repository}. 
Directly incorporating knowledge from these paths into the LLM would significantly distort its reasoning and judgment. 
However, these noisy paths can be easily identified and excluded by considering and analysing the input question.
For example, we can exclude the noisy path (\uppercase\expandafter{\romannumeral1}) by considering that the question is irrelevant to "river". 
In addition, the noisy path (\uppercase\expandafter{\romannumeral2}) can be excluded by taking into account the semantic meaning more deeply such as a specific restriction phrase, "where children keep their assets".

Building on the insights from above analyses, this paper introduces a question-guided knowledge graph re-scoring method (\textbf{Q-KGR}) to eliminate the negative effect of noisy pathways for specific questions, thereby focusing attention on precise factual knowledge.
Specifically, we develop an edge re-scorer to assign a relevance score to each edge by computing the semantic similarity between the specific question and each edge triplet to obtain a re-scored knowledge graph.
% Based on this, the pathway score is calculated by multiplying the scores of each edge within the corresponding pathway.
% Based on this, we use a bilinear layer to estimate the probability distribution of edge presence and design L-layers GNNs to model structured knowledge.  
And then we extend the graph attention network to incorporate edge scores during feature aggregation.
In addition, we design \textbf{Knowformer}, a customized transformer where a knowledge projector is utilized to align and inject the information of knowledge graphs into the standard transformer architecture, thereby enhancing its reasoning capabilities.

Furthermore, Knowformer can naturally adapt to LoRA \cite{lora}, and the combined use can effectively achieve task adaptation and knowledge injection. 
We conduct extensive experiments on four KGQA benchmarks, and the experimental results demonstrate that our method successfully eliminate noisy pathways and inject knowledge into LLM to perform better factual reasoning.
Our main contributions are summarized as follows:

%后面需要讨论
\begin{itemize}
    \item We propose Q-KGR to re-score the edges of knowledge graph and introduce Knowformer to align and inject structured knowledge into  LLM, thereby assisting it in conducting accurate factual reasoning.
    \item Extensive experiments on multiple KGQA benchmarks demonstrate the superiority of our method across different LLM architectures, sizes and setups.
    \item Further experimental analysis indicates that our approach effectively guides the LLM to concentrate on accurate factual knowledge and achieves parameterized knowledge alignment and injection.
\end{itemize}

\section{Related Work}
\paragraph{Knowledge Graph Question Answering}
Knowledge Graph Question Answering (KGQA) focuses on enabling machines to retrieve and utilize relevant knowledge from a KG (\citealp{cpnet}; \citealp{freebase}). This task requires models to excel in capturing accurate factual knowledge and complex reasoning abilities.

Capturing accurate factual knowledge requires maximizing recall during retrieval, followed by meticulous screening to filter out irrelevant information. Many methods have embarked on this journey (\citealp{unik1}; \citealp{unik2}; \citealp{graph_reasoning}). \citealt{kagnet} and \citealt{greaselm} propose to heuristically retrieve a subgraph from the KG for reasoning over KG. Similar to our method, \citealt{qagnn} introduces node relevance scores to filter irrelevant nodes. However, edges including subject, relation and object in KG capture richer information about interactions between entities. By comparision, using edge scores can provide more comprehensive views which in turn enhances the model's ability to accurately capture and utilize relevant pathways for more precise and meaningful answers. Besides, \citealt{great} and \citealt{counter} indicate that node relevance scores brings weak or even redundant effects in model optimization and our subsequent experiments also validate prior conclusions. On the other hand, because LLMs have demonstrated remarkable reasoning abilities on various tasks (\citealp{llm1}; \citealp{benchmark}; \citealp{llm2}; \citealp{dual}; \citealp{translation1}), many methods apply them to conduct reasoning tasks on KG (\citealp{rog}; \citealp{knowgpt}; \citealp{gnp}). Our method simultaneously achieves capturing accurate factual knowledge and evoking complex reasoning skills.

\paragraph{Knowledge Editing and Injecting for Large Language Model}
In the field of knowledge editing, many works have explored how factual knowledge is stored in LLM \cite{rome,edit2,edit3}. 
Through causal analysis and quantitative experiments, \citealp{ffn1} and \citealp{ffn2} demonstrates that parameterized knowledge in LLM is stored as key-value memory pairs in the feed forward Neural Network (FFN) of the transformer. 
Building on this, \citealp{kformer} is the first to attempt injecting text knowledge into parameter space of language models (LM). 
\citealp{calibrating} constructed an external knowledge vector database to calibrate factual knowledge in LLM. 
By comparison, we utilize GNNs to model and inject structured knowledge into LLMs in a layer-wised manner.

 \section{Preliminary}
\paragraph{Knowledge Graph}
We define knowledge graph (KG) as a multi-relation graph $\mathcal{G}=(\mathcal{V},\mathcal{E})$ where $\mathcal{V}$ is the set of entity nodes in the KG, $\mathcal{E} \subseteq \mathcal{V} \times \mathcal{R} \times \mathcal{V}$ is the set of edges that connect nodes in $\mathcal{V}$ and $\mathcal{R}$ represents the set of relaton types in the KG. 

\paragraph{Knowledge Retrieval}
Given a question $q$ and all candidate answers $\mathcal{A}_q$, we follow the previous work \cite{kagnet} to associate the entities mentioned in the question and all candidate answers from $\mathcal{G}$. We use $\mathcal{V}_{qa}:= \mathcal{V}_q \cup \mathcal{V}_a$ to represent all the entities that appear in the question and answers, named topic entities. Then, we retrieve inner node entities $\mathcal{V}_{in}$ and corresponding edges $\mathcal{E}_{in}$ from KG on the $k$-hop paths among all topic entities. 

\paragraph{Knowledge Graph Question Answering (KGQA)} is a typical reasoning task based KG. The task aims to design a function $f$ to predict correct answer $a \in \mathcal{A}_q$ based knowledge from $\mathcal{G}$ i.e., $a=f(q,\mathcal{G})$.

\section{Methodology}

\begin{figure*}[t]
    \centering
    \includegraphics[page=1, scale=0.75]{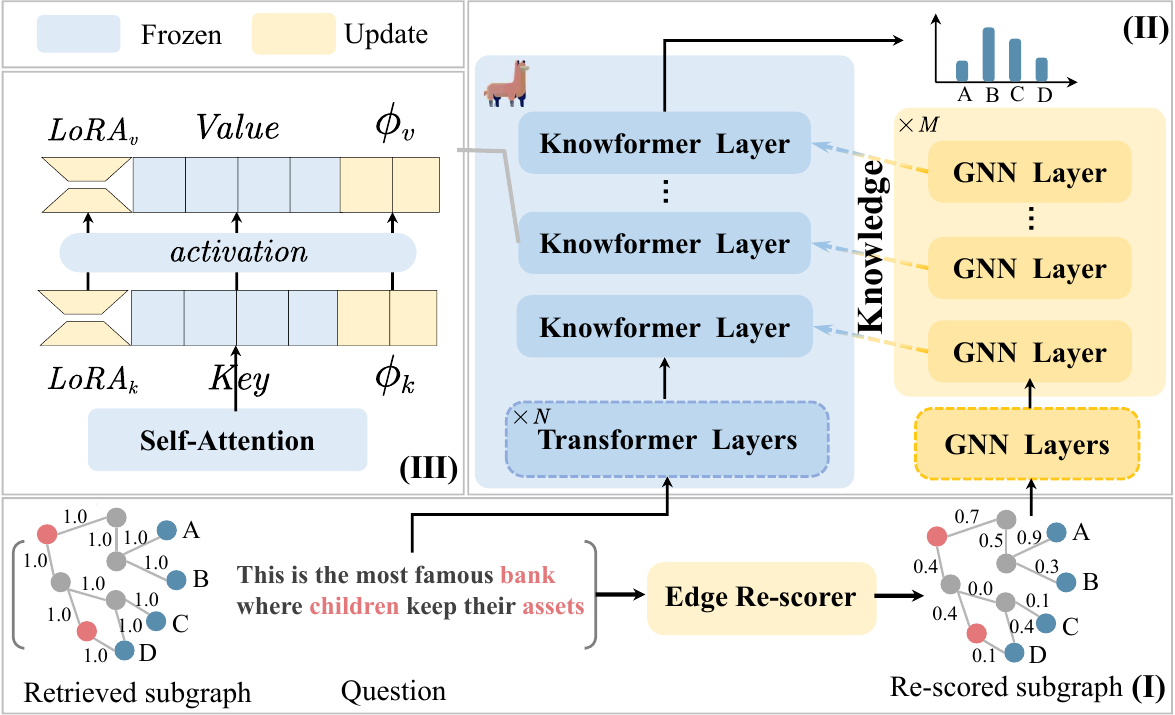}
    \caption{The overall framework of our method. Given a retrieved subgraph from origin KG, we revise it through question-guided knowledge graph re-scoring (Q-KGR, \uppercase\expandafter{\romannumeral1}). Then we utilize question text and revised subgraph to conduct knowledge modeling, injection and reasoning (\uppercase\expandafter{\romannumeral2}). Knowformer (\uppercase\expandafter{\romannumeral3}) consists of self-attention layers and a customized feed-forward network (FFN) layer. In the FFN layer, $key$ and $value$ represent original  weight matrices, $LoRA_k$ and $LoRA_v$ correspond to LoRA weight parameters, and $\phi_k$ and $\phi_v$ are the knowledge vectors mapped from graph latent space to the parameter space of FFN. Gray denotes frozen model parameters, while green indicates updated model parameters.}
    \label{fig:frame}
\end{figure*}

The overall framework of the proposed method is illustrated in Figure \ref{fig:frame}. 
Firstly, we employ an edge re-scorer to assign a relevance score to each edge based on its relevance to the question ($\S \ref{sec:reconstruction}$).
Additionally, we modify GAT to encode the re-scored subgraph, obtaining structured knowledge representations at each layer ($\S \ref{sec:gat}$). Furthermore, we develop a layer-wised knowledge injection method to inject knowledge from last $M$ layers in GNNs into corresponding Knowformer layers ($\S \ref{sec:inject}$). 
Finally, we process the question text sequentially through $N$ layers of the transformer and $M$ layers of Knowformer to predict the probability distribution of the candidate answers.

\subsection{Question-guided Knowledge Graph Re-scoring}
\label{sec:reconstruction}
For eliminating the noisy pathways, we introduce a question-guided edge re-scorer to assign each triplet a relevance score by computing the semantic similarity between the specific question and each edge.
Specifically, we sample each node pair (subject and object) from the subgraph and feed them into a pre-trained language model (PLM) to gain the embeddings as $e_s$ and $ e_o$:
\begin{equation}
    e_s, e_o=PLM(subject,object).
\end{equation}
To leverage the core semantic information in question $q$, we extract entity representations from encoded question by the same PLM and aggregate them through an average pooling layer as $e_q$. 
Given the computational constraints posed by the large number of edges in the original graph, we keep the parameters of the PLM frozen to ensure efficient processing. 
Finally, we utilize a bilinear layer to estimate the relevance score for each edge as: 
\begin{equation}
    \eta = Normalize (Bilinear([e_s, e_o], e_q)),
\end{equation}
Regarding normalization, we investigate two strategies: (1) Gumbel-max \cite{gumbel}: modeling predicted score for an edge as a hard (one-hot) label, $\eta \in \mathbb{R}^{2 \times 1}$ indicating whether the edge exits. (2) Gumbel-softmax \cite{gumbel}: modeling the score as a soft label, representing the score of relevance and irrelevance.
Since there is no gold graph structure as supervised signal to train the proposed edge re-scorer individually, we incorporate the predicted score as a latent variable to train an end-to-end model. Moreover, to address the issue of gradient propagation when using hard labels, we further refine the one-hot label as follows:
\begin{equation}
    \eta = \eta + \mathrm{one\_hot} (\eta) -sg (\eta)
\end{equation}
where $sg$ stands for the "stop gradient" operator (also called .detach() in PyTorch) which does not affect the forward pass, but in the backward pass it does not back-propagate the gradient to its input, meaning it is treated as a constant.
We find that both strategies are effective, but in the main experiments, we use the soft label.

\subsection{Re-scored Knowledge Graph Modeling}
In order to better aggregate and extract extensive knowledge from knowledge graph, we first expand the subgraph by adding a \textbf{\textit{global node}} as $\mathcal{V}_g$ connecting to all topic nodes.
\label{sec:gat}
Based on this, we extend the graph attention network (GAT)~\cite{gat} to encode the re-scored knowledge graph, which induces node representations via iterative message passing between neighbors on the graph. 

In the $l+1$ layer, we update the representation $h_t^{l+1} \in \mathbb{R}^{D}$ of each node $\mathcal{V}_{t} $ by 
\begin{equation}
    h_{t}^{\left( l+1 \right)}=f_m\left( \sum_{s\in N_t\cup \left\{ t \right\}}{\alpha _{st}m_{st}} \right) +h_{t}^{\left( l \right)}
\end{equation}
where $f_m: \mathbb{R}^D \rightarrow \mathbb{R}^D$ is a two-layer MLP, $N_t$ denotes the neighbors of node $\mathcal{V}_{t}$, $m_{st} \in \mathbb{R}^D$ and $\alpha_{st}$ represent the information and weight from each neighbor node $\mathcal{V}_{s}$ to node $\mathcal{V}_{t}$. And $\alpha_{st}$ is calculated as: 
\begin{equation}
    \alpha_{st} = \frac{\eta \cdot \exp(\gamma_{st})}{\sum_{s' \in N_t \cup \{t\}} \exp(\gamma_{s't})},\gamma_{st} = \frac{q_{s}^{T}k_t}{\sqrt{D}}
\end{equation}
where $q_s$ and $k_t$ correspond to the query and key in the GAT, obtained from node and relation type embeddings through multiple MLPs \footnote{The calculation method for $m_{st}$,$q_s $ and $k_t$ is based on \citealt{qagnn}.} and we use edge relevance score \(\eta\) from Section \ref{sec:reconstruction} to scale the original attention weights. For each node, we use the same PLM as section \ref{sec:reconstruction} to initialize the representation, and we initialize the global node as 0. 

Finally we obtain the embeddings of all nodes in each layer and for last $M$ layers, we use the representation of the global node, $h_l$ as external graph knowledge. 
We will align and inject them into LLMs in section \ref{sec:inject}.

\begin{table*}[t]
\centering
\scalebox{0.95}{
\begin{tabular}{llcccccc}
\toprule
\textbf{LLM} & \textbf{Method}& \textbf{OBQA}& \textbf{Riddle}& \textbf{ARC}  &\textbf{PIQA}&\textbf{Average}\\
 \midrule
\multirow{9}{*}{\textbf{FLAN-T5-XL}}
&LLM-only &69.20&53.73 &68.24 &58.43 & 62.24\\
&REL &61.80&43.33&64.12&57.56&57.71\\
% &KG Flattening BFS &62.80&44.12&63.86\\
&KAPING TH &58.80&40.78&63.52&52.34&53.86\\
&KAPING OH &60.00&41.37&63.09&51.69&54.04\\
&Prompt Tuning &72.20&53.33&70.64&60.83 &64.25\\
&Full Fine-tuning &82.8&74.12&73.30& 63.55&73.44\\
&LoRA &80.40&72.94&71.33& 63.76&72.11\\
&LoRA+GNP&83.40&75.49& 72.45& 64.31&73.91\\
&LoRA+Ours &\textbf{85.00} &\textbf{78.43}&\textbf{73.38}&\textbf{87.03} &\textbf{80.96} \\
 \midrule
\multirow{9}{*}{\textbf{FLAN-T5-XXL}}
&LLM-only &76.8 &61.37&68.93& 56.58&65.92\\
&REL &72.80&53.53&66.78&56.80&62.48\\
&KAPING TH &60.60&48.43&57.25&53.21&54.87\\
&KAPING OH &60.00&47.65&56.65&51.69&54.00\\
&Prompt Tuning &78.80&61.37&74.85&61.26 &69.07\\
&Full Fine-tuning &89.40&80.78&76.82& 65.61&78.15\\
&LoRA &88.60&74.90&78.54& 65.61&76.91\\
&LoRA+GNP&89.60&76.67&78.71& 65.94&77.73\\
&LoRA+Ours &\textbf{90.00}&\textbf{80.98}&\textbf{79.78}&\textbf{90.08} & \textbf{85.21} \\
\bottomrule 
\end{tabular}
}
\caption{\label{tab:main}
Overall experimental results on commonsense reasoning tasks. The best results across different LLM sizes and setups are highlighted in bold. Accuracy is used as the evaluation metric. 
}
\vspace{-0.5cm}
\end{table*}

\subsection{Knowformer}
\label{sec:inject}
In a LLM, we use $M$ Knowformer layers to replace the original Transformer layers that are closest to the LLM’s final output for structured knowledge injection into LLM to assist in conducting factual reasoning. 
For a single Knowformer layer, it inherits the parameters of the original Transformer layer at the corresponding position. 
A standard transformer consists of a multi-head self-attention layer and a feed-forward network (FFN) layer, typically comprising two or three linear layers with an activation function. 
Supposing that the input of FFN is $x$ and the output of FFN is computed as:
\begin{equation}
    FFN(x)=(x \cdot key )\cdot value
\end{equation}
where \textit{key, value} $\in \mathbb{R}^{d_m \times d}$ are parameter matrices of the first and the second layer of FFN, respectively.\footnote{We simplify  the FFN layer by omitting the activation layer and bias and take the FFN composed of two linear layers as an example.}
As shown in Figure \ref{fig:frame}, the Knowformer modifies the FFN by coupling $key, value$ with aligned structured knowledge, $\phi_k,\phi_v$. 
The computation within the modified FFN can be described as follows:
\begin{equation}
FFN(x)=(x \cdot [key:\phi_k]) \cdot [value:\phi_v].
\end{equation}

To inject the structured knowledge, we first obtain the structured knowledge representation $h \in \mathbb{R}^{1\times D}$ for a certain layer from section \ref{sec:gat}.
Subsequently, we align it with parameterized knowledge in FFN using two linear projectors: 
\begin{equation}
\phi_{k}=Pr_k h, \phi_{v}=Pr_v h,
\end{equation}
where $Pr_k, Pr_v: \mathbb{R}^{D} \rightarrow \mathbb{R}^{d}$ are two parameter matrices and $d$ is the input dimension of FFN. 
Inspired by \citealt{lora}, we use random Gassian distribution and zero to initialize $Pr_k$ and $Pr_v$ respectively.
It should be noted that our method is highly flexible, seamlessly integrating with LoRA, and its combined use effectively facilitates task adaptation, knowledge alignment, and injection. 
Finally, our method is architecture-agnostic. As shown in the Figure \ref{fig:frame}, we illustrate general Transformer layers and do not distinguish architecture of LLM e.g., encoder-decoder or decoder-only. Our knowledge injection is performed in the several layers that are nearest to the LLM’s final output.
\section{Experiments}
In this section, we conduct extensive experiments on four KGQA benchmarks including OpenBookQA (OBQA)~\cite{obqa}, AI2 Reasoning Challenge (ARC)~\cite{arc}, RiddleSense (Riddle)~\cite{riddle} and Physical Interaction Question Answering (PIQA)~\cite{piqa} to compare the performances of different methods. We use ConceptNet~\cite{cpnet}, a general-domain knowledge graph as our structured knowledge source $\mathcal{G}$. A detailed explanation of the datasets and the knowledge graph is provided in the appendix \ref{sec:experiment}.

\subsection{Baselines}
We use encoder-decoder architecture LLM, FLAN-T5-XL (3B) and FLAN-T5-XXL (11B) \cite{flan} as the backbone to conduct experiments. We compare eight baselines including \textbf{LLM-only}, KG Flattening that flattens the nodes in the graph into a sequence via relevance score (\textbf{REL}) ranking \cite{rel}, \textbf{KAPING} \cite{kaping} that injects the important KG triples within one-hop (OH) and two-hop (TH) neighborhoods, \textbf{Prompt Tuning} \cite{prompt_tuning} that regards embedding knowledge (soft prompts) as prefix of the text question after embedding layer in LLM, \textbf{LoRA} \cite{lora} that updates partial LLM parameters, \textbf{Full Fine-Tuning} and \textbf{GNP} \cite{gnp} which uses GNN to model retrieved subgraph as soft prompts to argument LLM. Besides, we also compare other baselines using encoder-only model (RoBERTa-Large)~\cite{roberta} and decoder-only model (LLaMA2-7B)~\citealt{llama2}. Appendix \ref{sec:experiment} provides more details on these baselines and experimental setups.

\begin{table}
\centering
\scalebox{0.85}{
\begin{tabular}{llllllll}
\toprule
\textbf{LM} & \textbf{Method}& \textbf{OBQA}& \\
 \midrule
\multirow{7}{*}{\textbf{RoBERTa-Large}}
&Full fine-tuning &64.80\\
&QAGNN&67.80 \\
&GreaseLM &66.90\\
&DRLK &70.20\\
&GSC &70.33 \\
&Rumination &70.30 \\
&Ours &\textbf{71.00} \\
 \midrule
\multirow{5}{*}{\textbf{LLaMA2-7B}}
&LLM-only & 31.80 \\
&Full Fine-tuning &74.00\\
&LoRA &71.80 \\
&LoRA-Ours &75.40 \\
&Full Fine-tuning-Ours &\textbf{77.00}\\
\bottomrule 
\end{tabular}
}
\caption{\label{tab:others}
Results on OBQA dataset using Encoder-only (RoBERTa-Large) and Decoder-only (LLaMA2-7B) architectures.
}
\vspace{-0.5cm}
\end{table}

\subsection{Main Results}
Table \ref{tab:main} compares the results of the proposed method with previous systems on four widely-used benchmark datasets. 
It can be firstly observed that our method with LoRA achieves the best results compared to other systems. In particular, our approach achieves a substantial breakthrough on the PIQA dataset, surpassing all baselines by a significant margin of absolute 20 percentage points.
Additionally, compared to the state-of-the-art system GNP, which uses a GNN to directly encode the retrieved subgraph as soft prompts and inject them into an LLM, our method obtains consistently better results. This achievement underscores the effectiveness of our approach in reconstructing the knowledge graph and injecting knowledge.

We further verify the generalization ability of the proposed method across both encoder-only and decoder-only architectures. 
Table \ref{tab:others} presents the performance of various systems using RoBERTa-Large and LLaMA2-7B as the underlying architectures. 
It is evident that our approach surpasses a series of strong baselines within the RoBERTa-Large model architecture. 
With the LLaMA2-7B model, our method, particularly when integrated with LoRA, exceeds full fine-tuning by 1.4 percentage points. 
When combined with full fine-tuning, our approach significantly outperforms other methods. 
Overall, our method has demonstrated exceptional performance across various model architectures, sizes, and training setups, confirming its effectiveness.

\begin{table}[t]
  \centering
  \renewcommand{\arraystretch}{1.1}
\scalebox{0.73}{
    \begin{tabular}{lccccc}
    \toprule
    
    \multirow{2}{*}{\textbf{Model}}& \multirow{2}{*}{\textbf{Variant}}& \multicolumn{2}{c}{\textbf{Riddle}} & \multicolumn{2}{c}{\textbf{ARC}} \\ 
    \cmidrule(l){3-4} \cmidrule(l){5-6}&
    &\textbf{dev}  & \textbf{test}  & \textbf{dev}  & \textbf{test} \\
    \midrule
    \multirow{7}{*}{\textbf{FLAN-T5-XL}}
 &w/o Q-KGR&72.99&73.73&77.26&72.61\\
 &NRS&72.73 &74.31&77.26 & 72.55 \\
  &HL&73.91&71.57&76.59&\textbf{73.81}\\
 % &+NRS+KGR&14.89 &4.15&20.63&4.87\\
 % &KGR(Ours)&14.89 &4.15&20.63&4.87\\
 % &+HL+KGR&14.89 &4.15&20.63&4.87\\
 &w/o PKI&73.02 & 72.94&74.21 &71.33 \\
 &SP& 72.45&76.30 &73.12 &72.56\\
 &w/o global &\textbf{74.95} & 74.31& 75.25&72.53 \\
 % &+PSI(Ours)&14.89 &4.15&20.63&4.87\\
&\textbf{Ours}&74.67&\textbf{78.43} &\textbf{77.59}&73.38 \\
\midrule
\multirow{7}{*}{\textbf{FLAN-T5-XXL}}
 &w/o Q-KGR&78.82  & 79.61&79.60 & 78.50 \\
 &NRS& 78.72&79.31&80.27 & 78.10\\
 &HL&\textbf{80.82}&\textbf{82.16}&79.60&78.75\\
 % &+NRS+KGR&14.89 &4.15&20.63&4.87\\
 % &KGR(Ours)&14.89 &4.15&20.63&4.87\\
 % &+HL+KGR&14.89 &4.15&20.63&4.87\\
  &w/o PKI&74.02 & 74.90&77.45 &78.54\\
 &SP&74.65 & 76.42&76.53 & 78.69\\
&w/o global &80.63 &79.61 &76.60 &76.35\\
 % &+PSI(Ours)&14.89 &4.15&20.63&4.87\\
&\textbf{Ours}&79.84 &80.98 & \textbf{80.27}&\textbf{79.78} \\
    \bottomrule
    \end{tabular}
    }
    \caption{Results of ablation study.}
  \label{tab:ablation}
% \vspace{-0.5cm}
\end{table}

\begin{table}[t]
\centering
\scalebox{0.95}{
\begin{tabular}{llcccccc}
\toprule
\textbf{Method}& \textbf{OBQA} &\textbf{CSQA}&\textbf{MedQA}\\
 \midrule
QAGNN &67.80&73.41&38.0\\
QAGNN-ours &68.56&75.02&38.49\\

\bottomrule 
\end{tabular}
}
\caption{\label{tab:qagnn}
Performances using node relevance scoring from \citealp{qagnn} and edge relevance re-scoring on the OBQA~\cite{obqa}, CSQA~\cite{csqa} and MedQA~\cite{medqa} dataset.
}
% \vspace{-0.5cm}
\end{table}

\begin{table*}[!ht]
    \centering
    \scalebox{0.73}{
    \begin{tabular}{llcccccc}
    \toprule
        LLM & Method & 0 & 10\%  & 20\% &  30\%  & 40\% & 50\%  \\ 
        \midrule
        \multirow{2}{*}{\textbf{FLAN-T5-XL}} 
        & w/o Q-KGR & 73.73 & 73.30 & 72.86 & 71.53 & 69.25 & 65.45  \\ 
        & Ours & 78.43 (+4.7) & 78.44 (+5.14)  & 78.30 (+5.44)  & 78.14 (+6.61) &  77.54 (+8.29)  & 76.28 (+10.83) \\ 
        \midrule
        \multirow{2}{*}{\textbf{FLAN-T5-XXL}} 
        & w/o Q-KGR & 79.61 & 79.53 & 78.70 & 76.01 & 74.31 & 72.27  \\ 
        & Ours & 80.98(+1.37) & 80.93(+1.40) & 80.56(+1.86) & 79.63 (+3.62) & 78.52 (+4.21) & 77.89 (+5.62)  \\ 
        \bottomrule
    \end{tabular}
    }
    \caption{Performances by adjusting the number of distractor nodes on the OBQA datasets.}
    \label{tab:noisy}
\end{table*}
\begin{figure}[t]
\includegraphics[width=\linewidth]{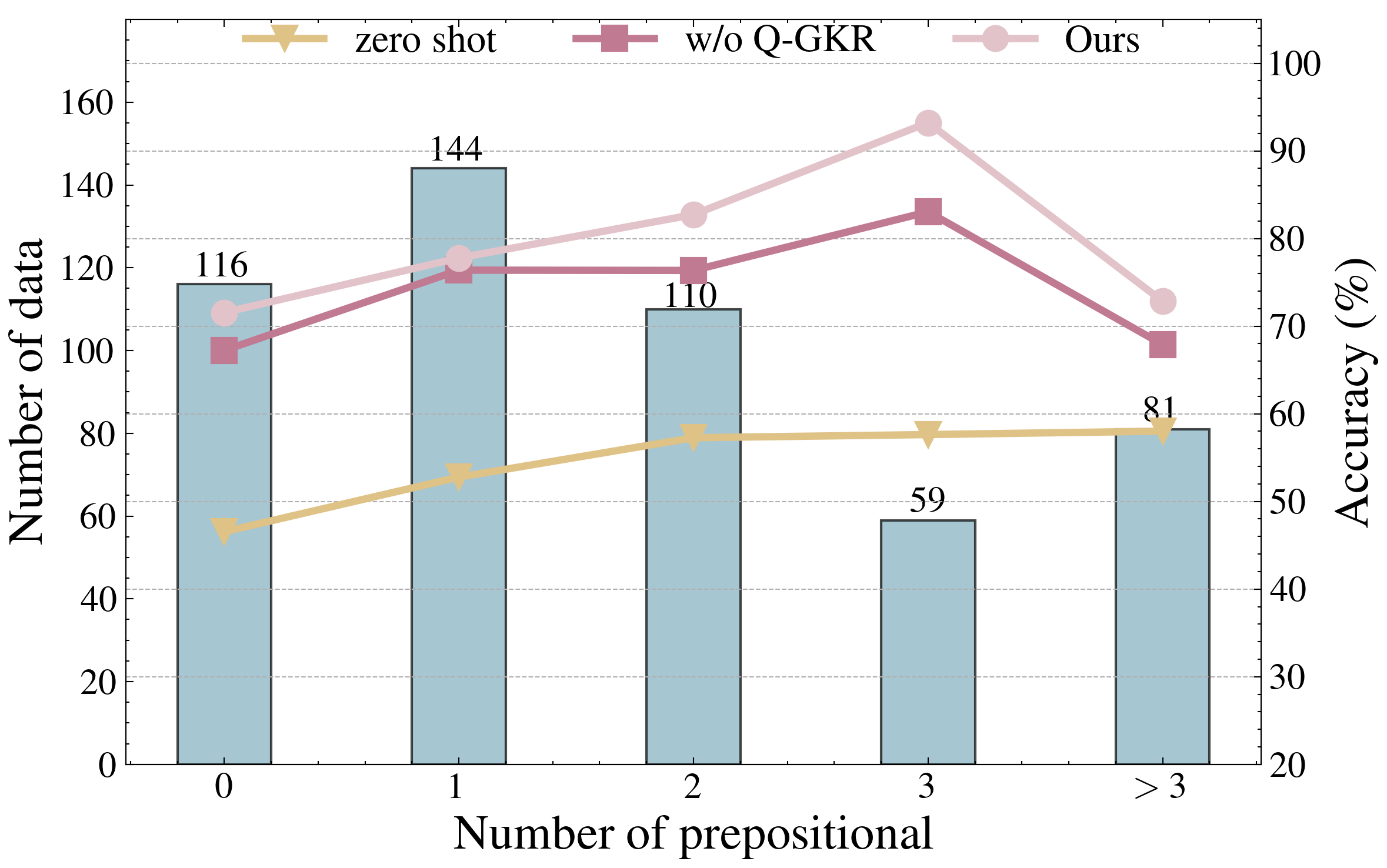}
\caption{Performance of different methods across divisions varying number of prepositional phrases.} 
\label{fig:prepositional}
\vspace{-0.5cm}
\end{figure}

\begin{figure*}[t]
    \centering
    \includegraphics[page=1, scale=0.6]{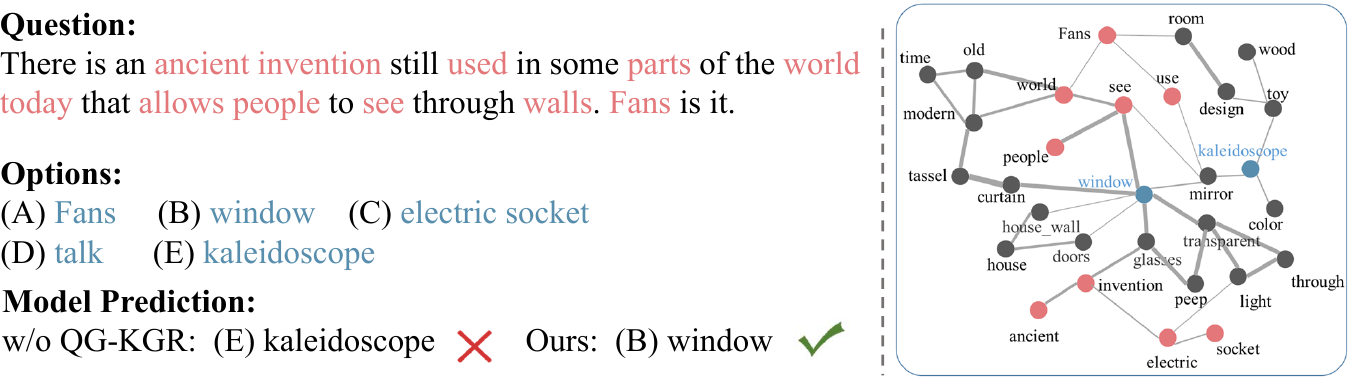}
    \caption{Qualitative analysis results.}
    \label{fig:case}
    \vspace{-0.5cm}
\end{figure*}

\subsection{Ablation Study}
We conduct extensive ablation experiments on Riddle and ARC datasets using FLAN-T5-XL and FLAN-T5-XXL to  investigate the effectiveness of three key components in our methods: 
(1) question-guided knowledge graph re-scoring (\textbf{Q-KGR});
(2) Re-scored  knowledge graph modeling (\textbf{KM});
(3) parmeterized knowledge injection (\textbf{PKI}). 

Targeted at Q-KGR, we design the following variants: 
removing Q-KGR (\textbf{w/o Q-KGR}); 
using node relevance scores from QAGNN~\cite{qagnn} to replace Q-KGR (\textbf{NRS});
utilizing hard labels from Section \ref{sec:reconstruction} for edge re-scoring (\textbf{HL}).
Targeted at KM, we compare our method with removing the global node and using the mean over all nodes representations as the graph representation (\textbf{w/o global}).
Targeted at PKI, we design the following variants: removing PKI (\textbf{w/o PKI});
representating the graph knowledge as soft prompt \cite{gnp} and inject it into embedding layer of LLM (\textbf{SP}). 

As illustrated in Table \ref{tab:ablation}, removing Q-KGR or replacing it with node scoring causes notable declines in the overall performance of the model, particularly on the Riddle dataset, highlighting the indispensable role played by Q-KGR. 
In addition, we compare the effects of using hard labels versus soft labels (notably, our method ultimately employs soft labels) as discussed in $\S$ \ref{sec:reconstruction} across various setups. 
It can be observed that soft labels generally leads to superior performance in the majority of cases.
Furthermore, when we remove the global node, there is a significant performance drop in the majority of experiments which indicates the crucial role of the global node in knowledge graph modeling at a global level.
Lastly, when we remove PKI or instead use SP to inject graph knowledge as soft prompts into the embedding layers of LLM \cite{gnp}, the overall results significantly declined compared to our method in all setups. 

To further compare Q-KGR with the node relevance scores (NRS) in QAGNN~\cite{qagnn}, we reproduce the method of QAGNN and replace NRS in QAGNN with our proposed edge relevance scoring to conducte experiments on the OBQA, CSQA \cite{kagnet}, and MedQA \cite{medqa} dataset in Table \ref{tab:qagnn}.
The results demonstrate that proposed method has obvious improvements compared to node relevance scoring in QAGNN and can better filter out retrieved irrelevant information.

\section{Analyses and Discussions}
We conduct comprehensive quantitative and qualitative analysis from various perspectives to validate the effectiveness of knowledge graph re-scoring and parameterized knowledge injection.

\subsection{Knowledge Graph Re-scoring}
\paragraph{Quantitative Analysis}

To show the denoising effectiveness of proposed method, we build a controllable setting where we gradually add the irrelevant nodes to the extracted subgraph and observe whether the performance is degrading.
Specifically, we limit the number of retrieved nodes to 200 and then collecte distractor nodes that are not in the 1-hop path among topic entities. 
We replace k\% of the nodes in the subgraph with distractor nodes, and then manually connect these distractor nodes to the topic nodes as a way to increase noise pathways. 
We test the performance of the proposed method and w/o Q-KGR on the Riddle dataset, varying k at 0, 10, 20, 30, 40, and 50, respectively. 
In Figure \ref{tab:noisy}, we observe that as k increases, the proposed method shows a greater improvements relative to w/o Q-KGR.
From the experimental results, we conclude that our method effectively reduces the noise in retrieving external knowledge, enabling more truthful factual reasoning.

To further validate whether the observed improvements indeed originates from filtering out irrelevant paths using edge re-scoring, we conduct another quantitative analysis.
Since there is no gold standard graph clue to measure the number of noisy pathways in the retrieved subgraph, we use the number of prepositional phrases as a proxy. 
For example, the question "What is that which follows you and be with you in the dark, but dies in the morning ?" contains three prepositional phrases: \textit{with you}, \textit{in the dark} and \textit{in the morning}. 
Each of these phrases provides additional search constraints and can introduce more pathways during knowledge retrieval.
Based on this, we divide the Riddle test dataset into five categories and use FLAN-T5-XL as the backbone to evaluate the performances of different methods. 
As shown Figure \ref{fig:prepositional}, when the number of prepositional phrases is 0 or 1, our method leads w/o QG-GKR by only a small margin. 
However, when the number of prepositional phrases exceeds 1, our method outperforms w/o QG-GKR obviously. 
This indicates that our method is an effective solution for addressing the challenges posed by noisy pathways.

\paragraph{Qualitative Analysis}
To intuitively understand the effectiveness of the proposed method, we present a case and visualize the relevance scores in Figure \ref{fig:case}, where the thickness of each edge represents the magnitude of the relevance score. 
It can be observed that our method correctly predicts the answer \textit{\textbf{window}}, with magnitude of the relevance scores between it and relevant nodes such as \textit{see}, \textit{transparent}, \textit{glasses} being significantly higher than the distractor option, \textbf{\textit{kaleidoscope}}. 
In addition, the thickness of the edge is strongly correlated with the semantic similarity of the edges to the question, thanks to the proposed edge re-scorer.
The above results demonstrate that our method indeed controls the flow of information on the graph based on the question, thereby eliminating noisy pathways and enabling the model to perform accurate factual reasoning. 
We provide more case analysis results in the Appendix \ref{sec:case}.

\begin{figure*}[t]
    \centering

    \begin{subfigure}{0.31\textwidth}
    \subcaption*{Epoch 1} 
        \includegraphics[width=\linewidth]{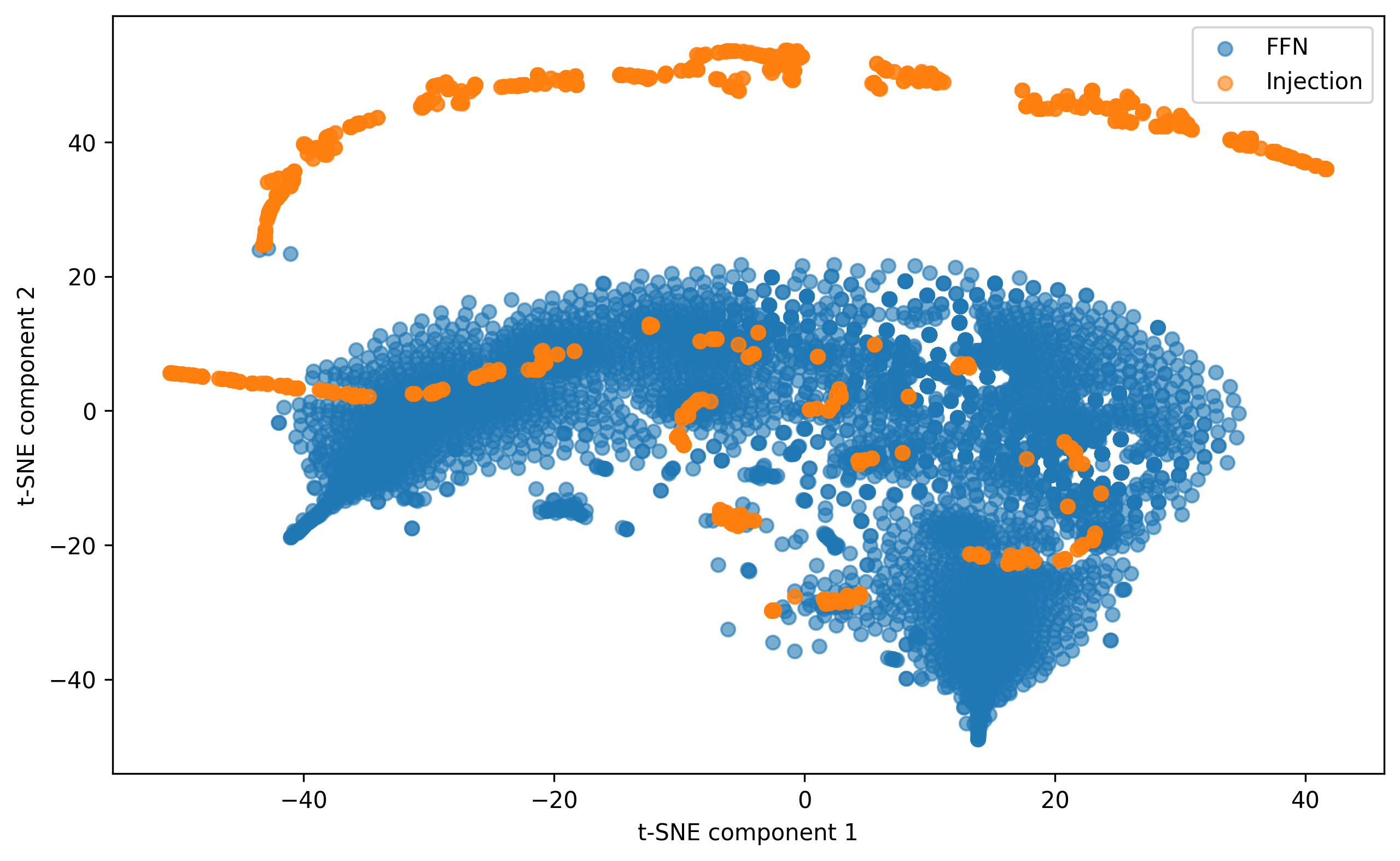}

    \end{subfigure}
    \hfill
    \begin{subfigure}{0.31\textwidth}
    \subcaption*{Epoch 3}
        \includegraphics[width=\linewidth]{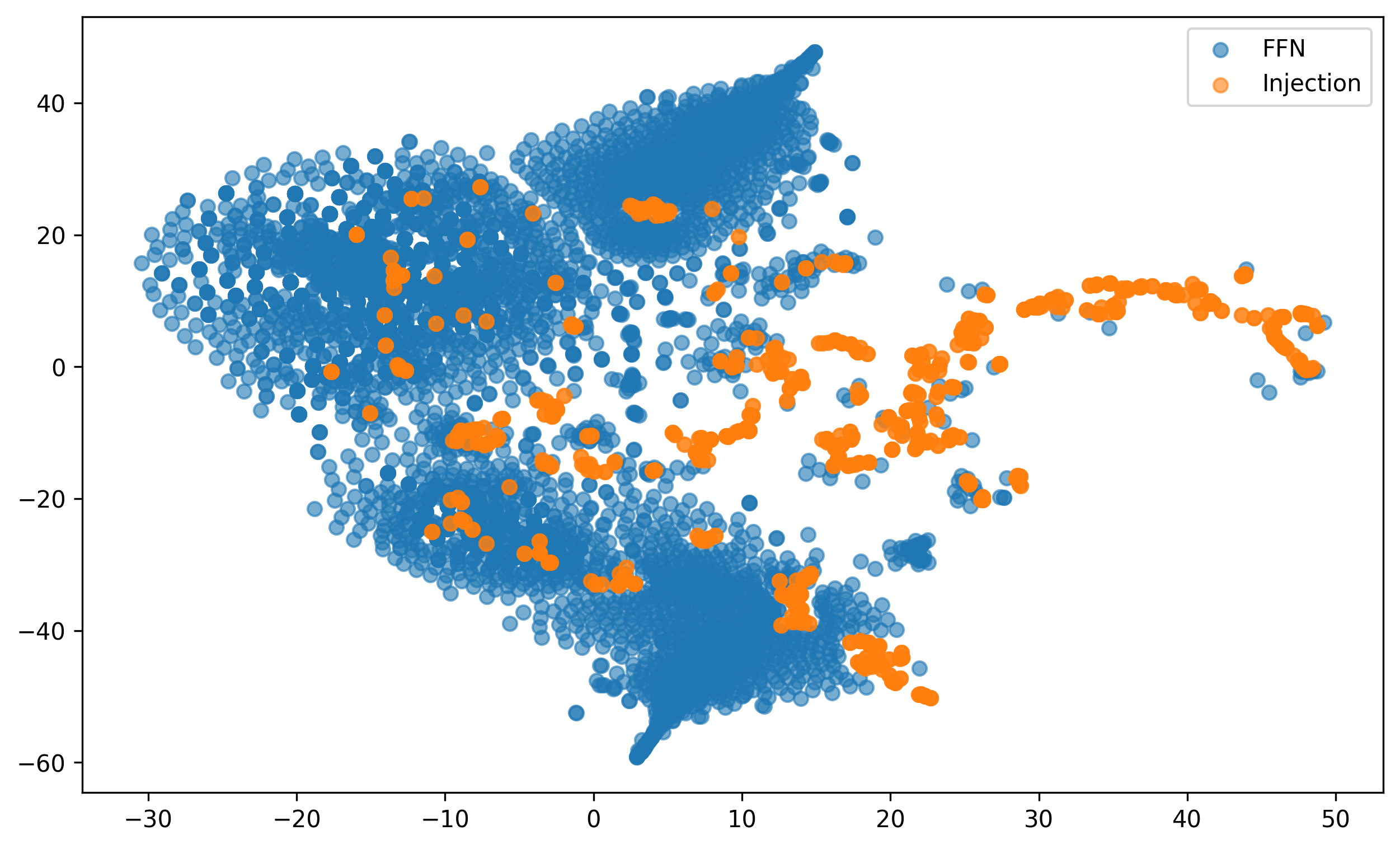}
        
    \end{subfigure}
    \hfill
    \begin{subfigure}{0.31\textwidth}
    \subcaption*{Epoch 5}
        \includegraphics[width=\linewidth]{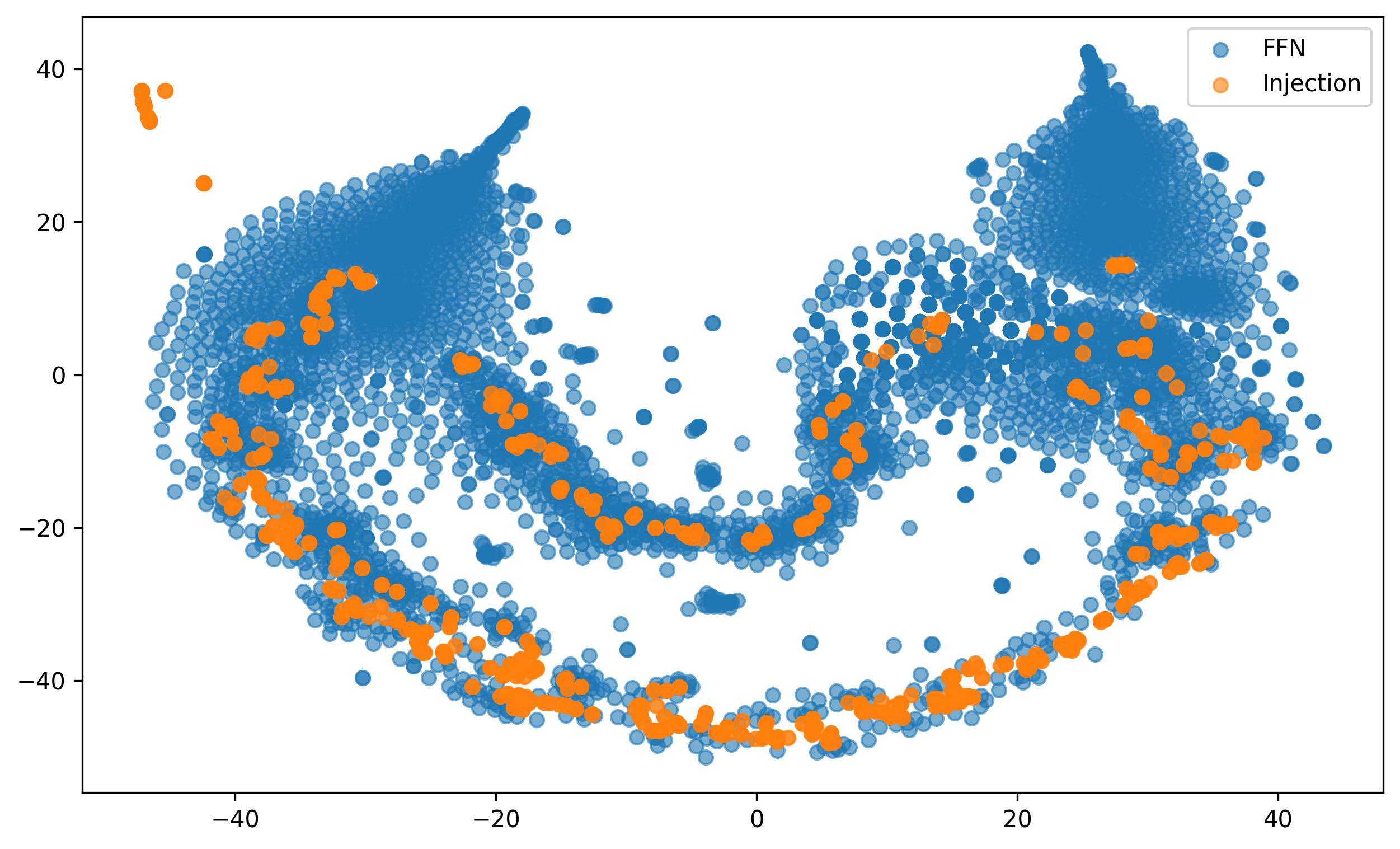}
        
    \end{subfigure}
    \caption{With the increase in training steps, it becomes apparent that the injected knowledge gradually integrates into the distribution of the FFN parameters.}
    \label{fig:step}
    \vspace{-0.5cm}
\end{figure*}

\begin{figure*}[t]
    \centering

    \begin{subfigure}{0.31\textwidth}
    \subcaption*{Layer 21} 
        \includegraphics[width=\linewidth]{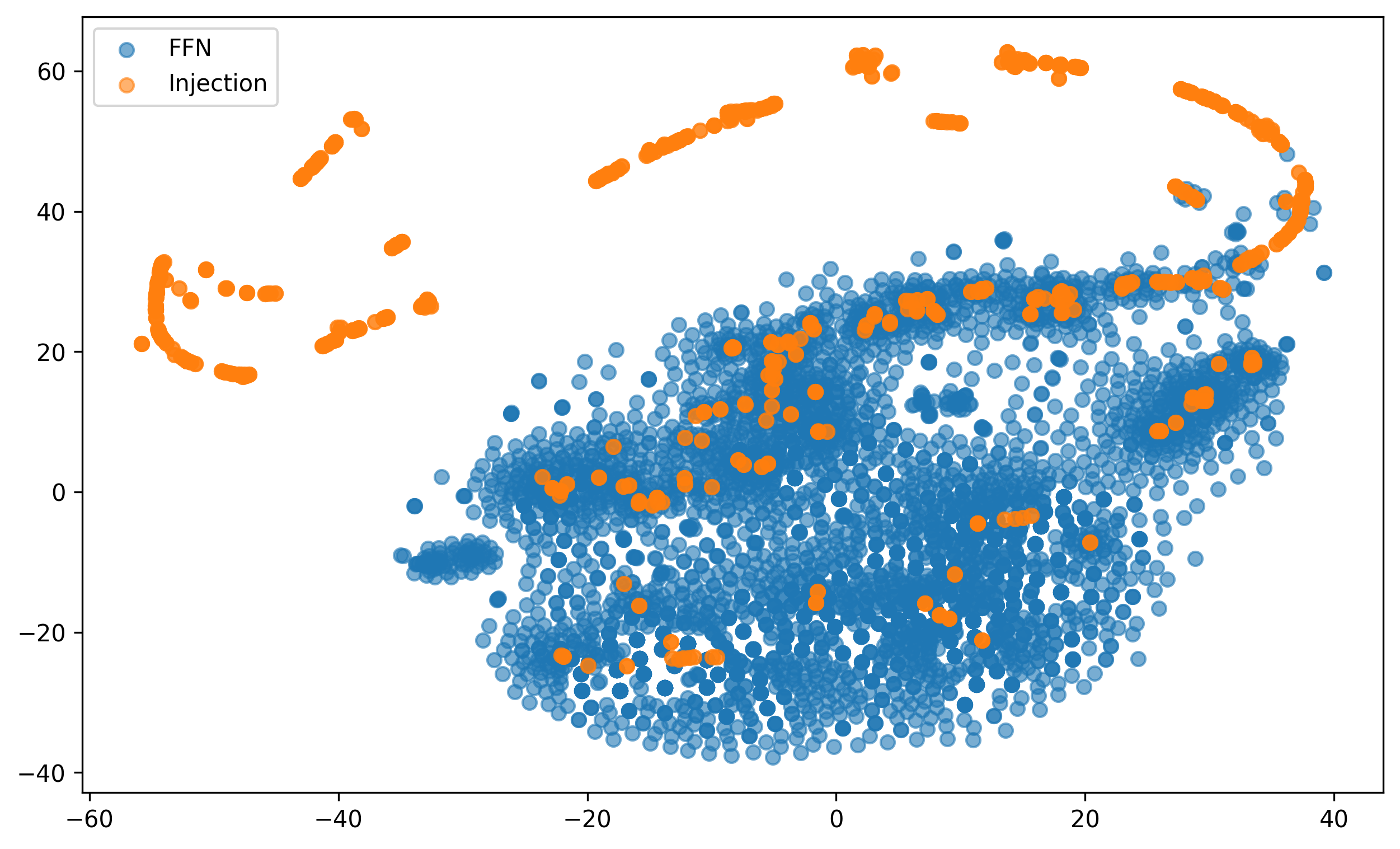}
        
    \end{subfigure}
    \hfill
    \begin{subfigure}{0.31\textwidth}
    \subcaption*{Layer 22}
        \includegraphics[width=\linewidth]{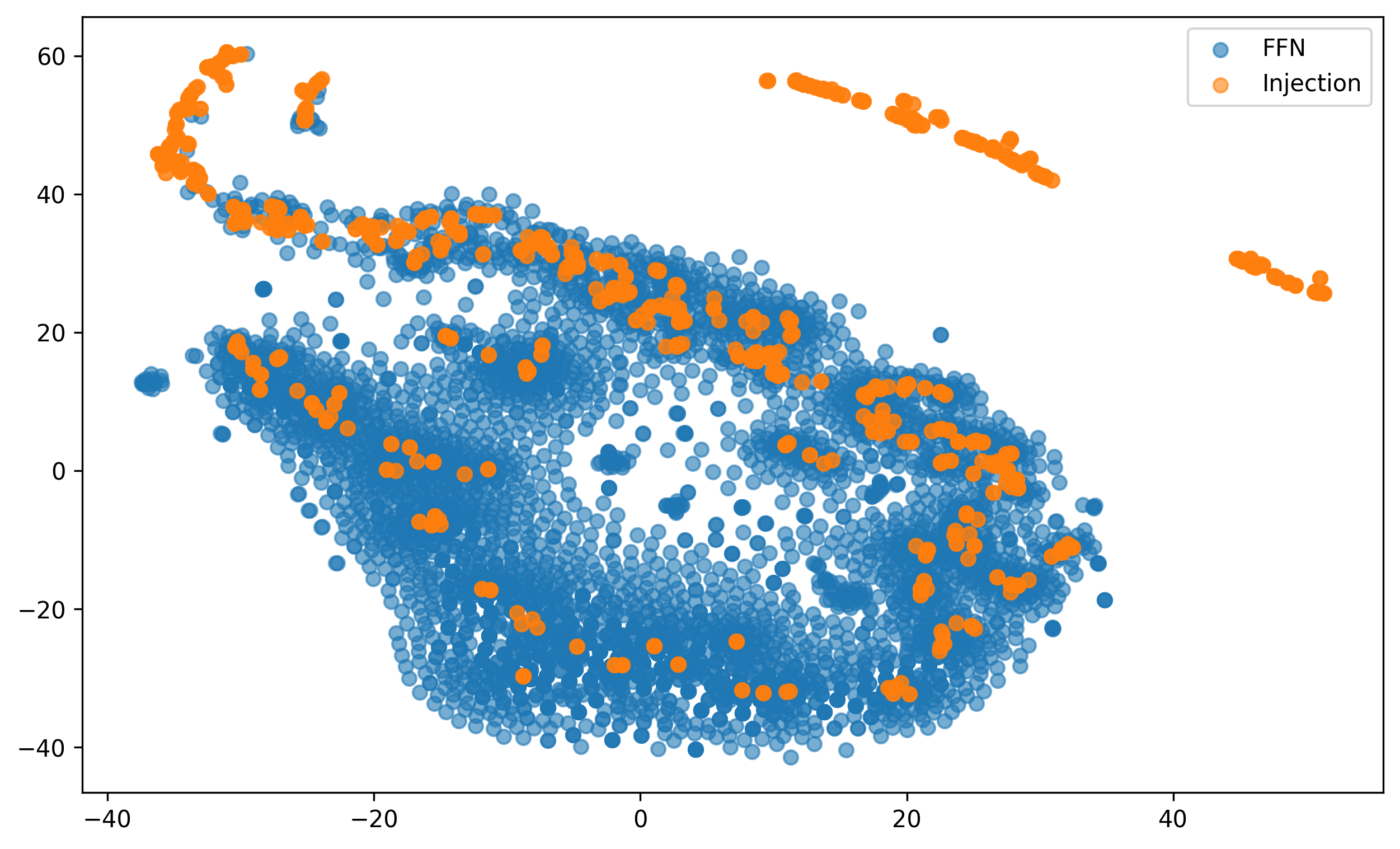}
        
    \end{subfigure}
    \hfill
    \begin{subfigure}{0.31\textwidth}
    \subcaption*{Layer 23}
        \includegraphics[width=\linewidth]{picture/all_question_ffn_2_layer_2.png}
        
    \end{subfigure}
    \caption{As the depth of the layers increases, the injected knowledge gradually integrates into the distribution of the FFN parameters.}
    \label{fig:layer}
    \vspace{-0.5cm}
\end{figure*}

\begin{figure}[t]
    \centering
    
    \begin{subfigure}{0.23\textwidth}
    \caption*{OBQA}
        \includegraphics[width=1\linewidth]{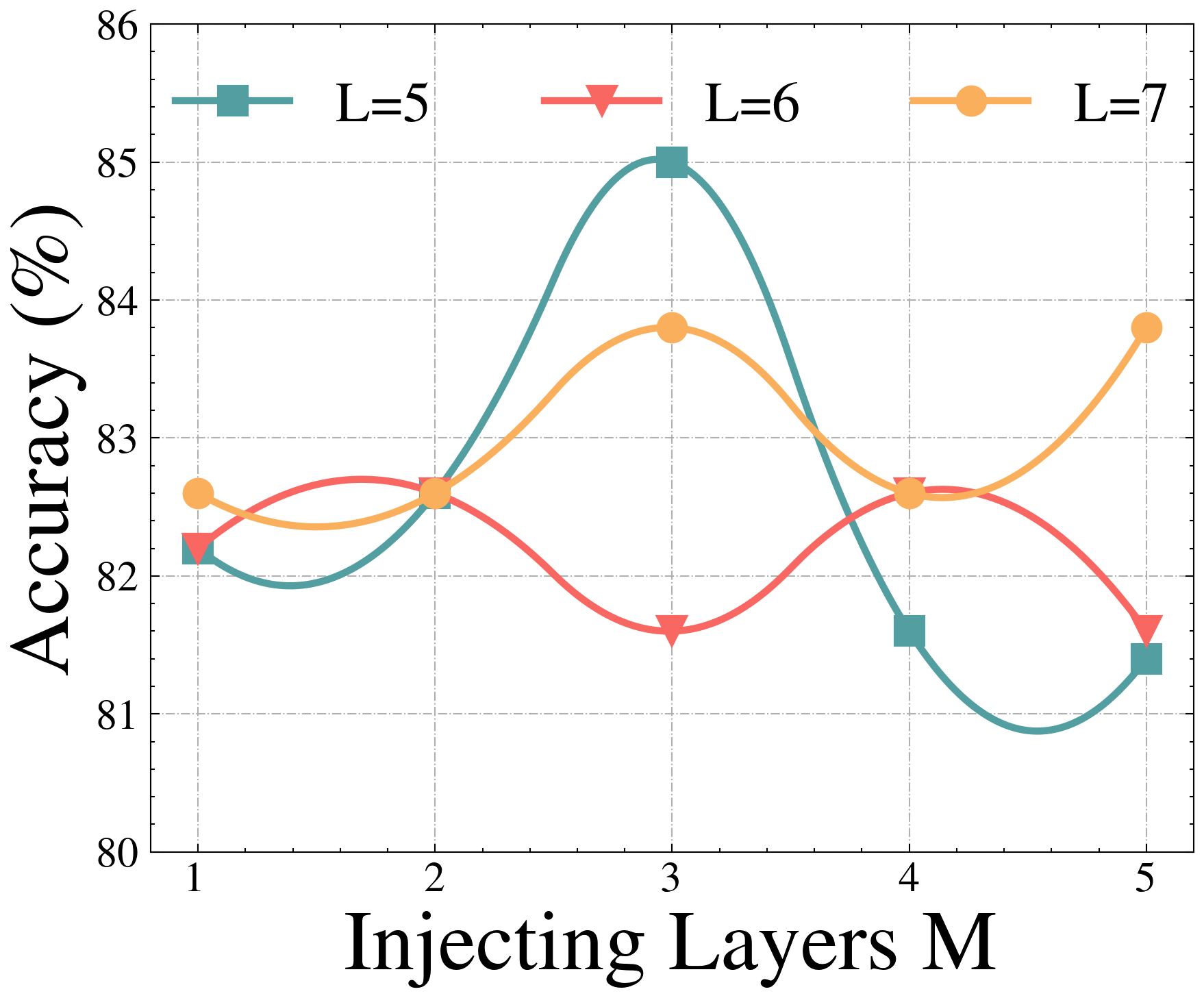}
        
    \end{subfigure}
    \hfill
    \begin{subfigure}{0.23\textwidth}
    \caption*{Riddle}
        \includegraphics[width=1\linewidth]{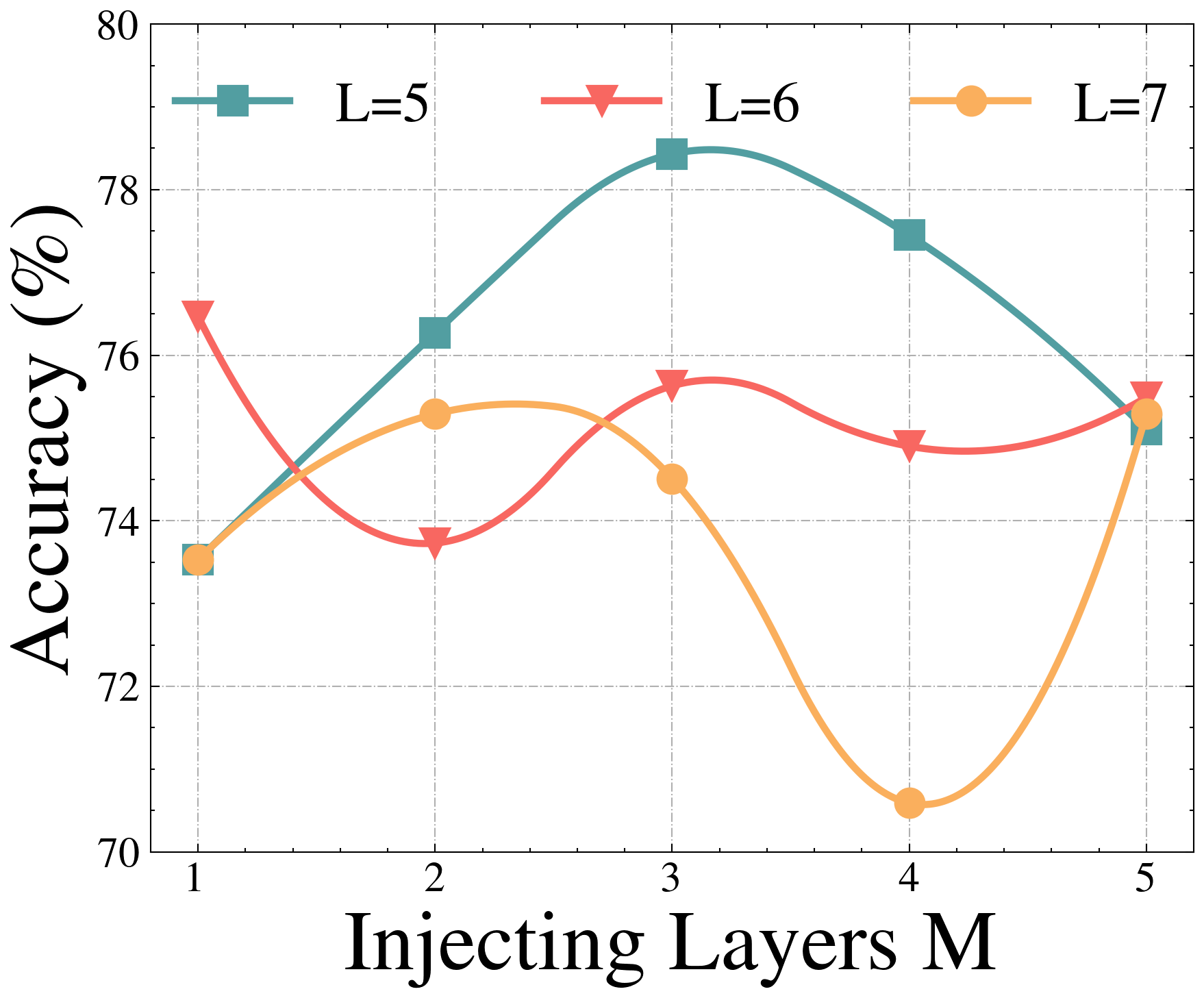}
        
    \end{subfigure}
    \caption{Performance by adjusting the layers of knowledge injection, as well as the number of layers in the GAT encoder on the Riddle and OBQA datasets.}
    \label{fig:hyper}
    \vspace{-0.5cm}
\end{figure}

\subsection{Parametered Knowledge Injection}
\paragraph{Knowledge Alignment}
To better understand the effectiveness of our method on knowledge alignment, we visualize the weight representation of the FFN and the inject knowledge (projected graph representation) using the t-SNE algorithm.
We train FLAN-T5-XL on the Riddle dataset for 5 epochs and select the models trained at epochs 1, 3, and 5 for analysis.
Specifically, we extract the knowledge vectors injected into the FFN and the corresponding FFN parameters and visualize them using the t-SNE algorithm \cite{t_sne}. 
Figure \ref{fig:step} displays the visualization results of the last layer (Layer 23) of different training steps. 
With the increase in training steps, it becomes apparent that the injected knowledge gradually integrates into the distribution of the FFN parameters, indicating the effectiveness of our training strategy for knowledge alignment.
Figure \ref{fig:layer} shows the visualization results of different layers after the last epoch of training. 
As the depth of the layer increases, the injected knowledge gradually integrates into the distribution of the FFN parameters, achieving the alignment of external knowledge. 
This may be because the LLM explores knowledge in its shallow layers and exploits this knowledge in its deeper layers by integrating more semantic information, similar to the conclusion in \cite{rome}.

\paragraph{Knowledge Injection}
We study the model performance when adjusting the layers of knowledge injection, as well as the number of layers in the GAT encoder.
As depicted in Figure \ref{fig:hyper}, as the number of injecting layers $M$ increases, our method generally shows an initial increase followed by a decrease in accuracy. Besides, it is observed that the total number of GNN layers $L$ should not be too large due to the limitation from graph smoothing \cite{smoothing}, ideally maintaining a balance around five layers.

\section{Conclusion}
In this paper, we propose a question-guided knowledge graph re-scoring method to eliminate noisy pathways for specific problem, directing attention towards precise factual knowledge. Furthermore, we design Knowformer, a customized transformer, through which we align and inject structured knowledge into the parameter space of LLM. Extensive experiments on multiple KGQA datasets across different LLM architectures, sizes and setups demonstrate the superiority of our method. And in-depth analysis experiments indictates that our model is particularly effective in handling complex questions characterized with significant noisy pathways in external knowledge. We also illustrate that our approach achieves effective knowledge alignment and injection for LLM.
\section{Limitation}

Our method improves the model's knowledge-based question answering by injecting commonsense knowledge but symbolic reasoning in Question Answering remains an open challenge.
From the technical perspective, we extract a representative node from a single layer of GNN and inject it into a Knowformer layer. However, incorporating more nodes intuitively might capture richer knowledge and potentially enhance model performance. Yet, this necessitates further validation regarding graph smoothing issues. Therefore, additional analysis is required to effectively address and balance this concern.

Additionally, aligning with human preferences ~\cite{preference,preference2} during the implementation of Retrieval-Augmented Generation (RAG) is another area that requires further exploration.
\section*{Acknowledgment}
We sincerely thank all the reviewers for their valuable and insightful comments, which have greatly enhanced the quality of our work. 
This work was supported by the National Natural Science Foundation of China (U23B2055 and 62276077), Guangdong Basic and Applied Basic Research Foundation (2024A1515011205), Shenzhen College Stability Support Plan under Grants (GXWD20220811170358002, GXWD20220817123150002 and GXWD2023-1130104007001).
The work of Zhao Kang was supported by Qiyuan Lab.

\bibliography{main}

\appendix
\section{Experiments}
\label{sec:experiment}
\subsection{Knowledge Graph and Datasets}
\paragraph{Knowledge Graph}
We conduct experiments in the general domain (commonsense reasoning) and we consider ConceptNet \cite{cpnet} that contains rich commonsense knowledge regarding the daily concepts. Node embeddings are initialized using the entity embeddings prepared by \citealt{kagnet}, which applies pre-trained LMs to all triples in ConceptNet and then obtains a pooled representation for each entity.
\paragraph{OpenBookQA} is a kind of question-answering dataset modeled after open book exams for assessing human understanding of a subject. It consists of 5,957 multiple-choice elementary-level science questions, which probe the understanding of a small “book” of 1,326 core science facts and the application of these facts to novel situations. Answering OpenBookQA questions requires additional broad common knowledge, not contained in the book. We use the original data splits in \citealt{knowledgeable}.
\begin{figure*}[h]
    \centering
    \includegraphics[page=1, scale=0.65]{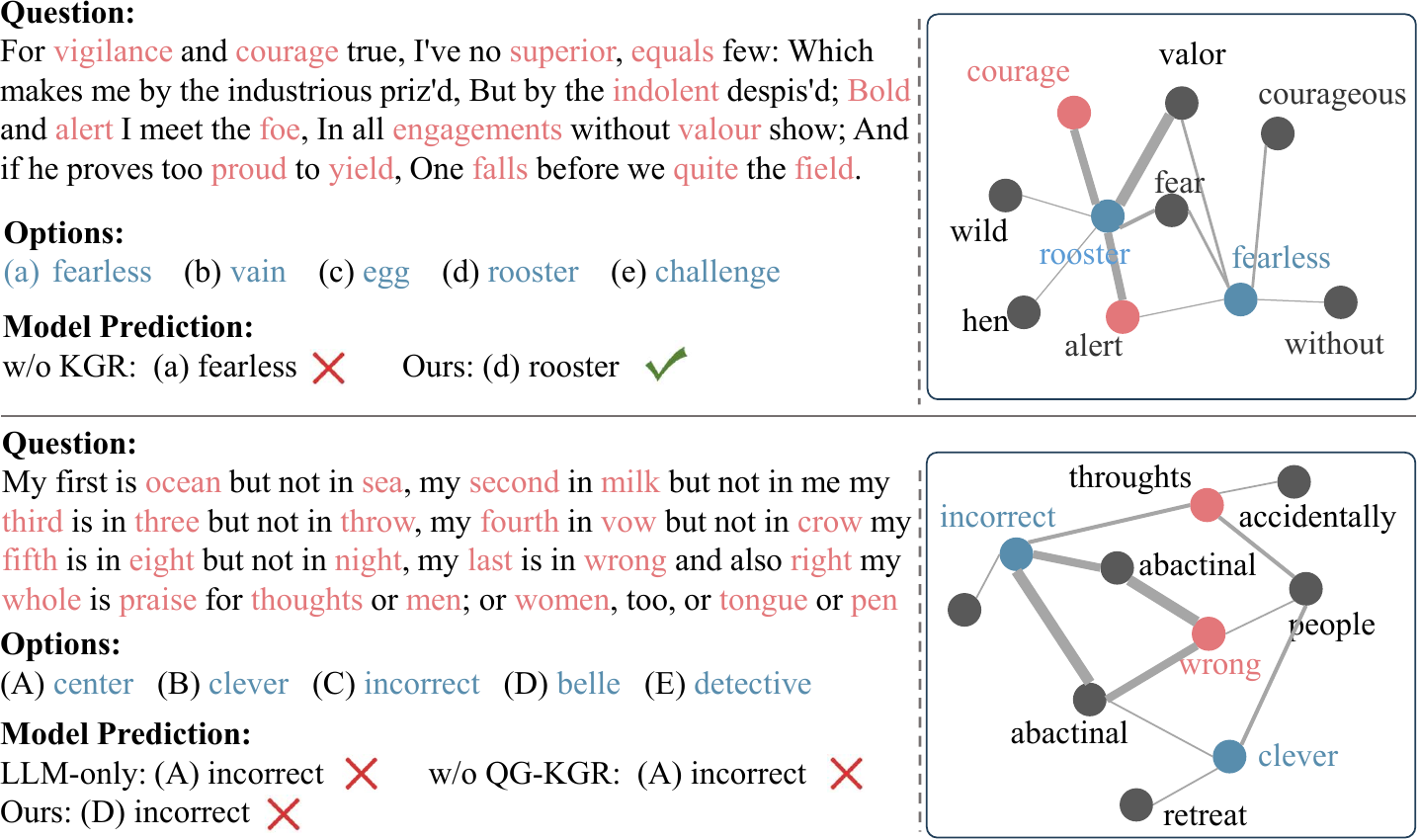}
    \caption{More case analysis results.}
    \label{fig:case_error}
\end{figure*}
\paragraph{ARC} consists of 7,787 science exam questions drawn from a variety of sources, including science questions provided under license by a research partner affiliated with AI2. These are text-only, English language exam questions that span several grade levels as indicated in the files. Each question has a multiple choice structure (typically 4 answer options). We evaluate challenge sets in it using the original data splits in \citealt{arc}.

\paragraph{Riddle} is a multiple-choice question answering task for answering riddle-style commonsense questions. It involves a challenging cognitive process, in which it requires complex commonsense reasoning abilities, an understanding of figurative language, and counterfactual reasoning skills, which are all important abilities for advanced natural language understanding (NLU). We split the dev set in half to make in-house dev/test sets in \cite{dragon}.
\paragraph{PIQA} is a dataset for commonsense reasoning, and was created to investigate the physical knowledge of existing models for naive physics reasoning focusing on how we interact with everyday objects in everyday situations.
\subsection{Implementation \& training details}
\label{sec:implementation}
For our method, we set the dimension of GNN module is 200 and number of layers of our GNN module from 5 to 7, with dropout rate~\cite{drop} 0.2 applied to each layer. We train the model with the RAdam \cite{radam} optimizer using A100-80G GPUs. We set the learning rate for LLM module from {1e-4, 3e-4}, learning rate for other modules as 1e-3, batch\_size from {16, 32}. The above hyperparameters are tuned on the development set. We choose FLAN-T5-XL (3B) and XXL (11B) ~\cite{flan} as the LLMs used in this paper. We adjust the maximum sequence length of LLMs to best fit the question length for each dataset. 

\subsection{Baseline}
In this setting of RoBERTa-Large, we compare with six baselines including \textbf{Full fine-tuning}, \textbf{QAGNN} \cite{qagnn} that proposes to use node relevance scores to filter irrelevant nodes and utilize joint training with GNN and pre-trained model (PLM), \textbf{GreaseLM} that designs a modality interaction unit to achieve information interaction between PLM and GNN, \textbf{drlk} \cite{drlk} that proposes a dynamic hierarchical reasoning method with PLM and knowledge graphs, \textbf{GSC} \cite{counter} simplifies the network structure and carries out some simple reasoning such as counting, \textbf{Rumination} that prompts PLM to generate inner knowledge and then inject it into PLM again.
In the setting of LLaMA2-7B, we compare with \textbf{LLM-only}, \textbf{Full fine-tuning} and \textbf{LoRA}.

\section{Case Study}
\label{sec:case}
In this section, we provide more results of case analysis in Figure \ref{fig:case_error}. In the fiour method successfully predicts the correct answer \textit{rooster} by accurately predicting the relevance scores between the question and all triplets in the knowledge graph and eliminating noisy pathways. In the second case, three methods failed to make the correct prediction. This is mainly because the question involves symbolic reasoning, and external commonsense knowledge injection could not offer substantial support which is a limitation of our method.

\end{document}